\documentclass{sig-alternate-05-2015}

\usepackage{graphicx}
\usepackage{epstopdf}
\usepackage{amsmath}
\usepackage{bbm}
\usepackage[font = footnotesize]{caption}
\usepackage[font = footnotesize]{subcaption}

\usepackage{array}
\usepackage{cite}
\usepackage{enumitem}
\usepackage{tikz}
\usetikzlibrary{arrows}
\usepackage{booktabs}
\usepackage{xifthen}
\usepackage{algorithm}
\usepackage{algorithmicx}
\usepackage[noend]{algpseudocode}
\usepackage[capitalize]{cleveref}
\usepackage{flushend}

\newcommand{\xxnote}[3]{}
\ifx\hidenotes\undefined
  \usepackage{color}
  \renewcommand{\xxnote}[3]{\color{#2}{#1: #3}}
\fi

\newcommand{\squishlisttwo}{
\begin{list}{$\bullet$}
{ \setlength{\itemsep}{0pt}
\setlength{\parsep}{0pt}
\setlength{\topsep}{0pt}
\setlength{\partopsep}{0pt}
\setlength{\leftmargin}{1em}
\setlength{\labelwidth}{1.5em}
\setlength{\labelsep}{0.5em} } }

\newcommand{\squishend}{
\end{list} }

\newtheorem{theorem}{Theorem}

\newtheorem{lemma}{Lemma}

\newcommand {\argmax}[1]{\underset{#1}{\operatorname{argmax}}}

\begin{document}


\title{Game-Theoretic Modeling of Human Adaptation \\ in Human-Robot Collaboration}

%

\numberofauthors{1}
\author{Stefanos Nikolaidis$^*$, Swaprava Nath$^\dagger$, Ariel D. Procaccia$^\dagger$, and Siddhartha Srinivasa$^*$\\
  \affaddr{\{$^*$Robotics Institute, $^\dagger$Computer Science Department\}, Carnegie Mellon University}\\
  \email{snikolai@cmu.edu}, \email{\{swapravn, arielpro\}@cs.cmu.edu}, \email{siddh@cmu.edu}
}


\CopyrightYear{2017} 
\setcopyright{acmcopyright}
\conferenceinfo{HRI '17,}{March 06-09, 2017, Vienna, Austria}
\isbn{978-1-4503-4336-7/17/03}\acmPrice{\$15.00}
\doi{http://dx.doi.org/10.1145/2909824.3020253}

\maketitle
\begin{abstract}
In human-robot teams, humans often start with an inaccurate model of the robot capabilities. As they interact with the robot, they infer the robot's capabilities and \textit{partially adapt} to the robot, i.e., they might change their actions based on the observed outcomes and the robot's actions, without replicating the robot's policy. We present a {\em game-theoretic model} of human partial adaptation to the robot, where the human responds to the robot's actions by maximizing a reward function that changes stochastically over time, capturing the evolution of their expectations of the robot's capabilities. The robot can then use this model to decide optimally between taking actions that reveal its capabilities to the human and taking the best action given the information that the human currently has. We prove that under certain observability assumptions, the optimal policy can be computed efficiently. We demonstrate through a human subject experiment that the proposed model significantly improves human-robot team performance, compared to policies that assume complete adaptation of the human to the robot.
\end{abstract}

\vspace{-2mm}
\section{Introduction}

A lot of work in robotics has focused on enabling robots to perform useful tasks for and with people. One of the main goals has been to make robots part of our everyday life, helping people as effective members of human-robot teams. In order to leverage recent advances in robot capabilities, human teammates should know what the robot can and cannot do: the robot's perceived capability should match its true capability. 

Prior work has shown that there is often a disconnect between users' perceptions and a robot's true capability, mainly due to lack of experience with working with robots and to the influence of popular culture~\cite{forlizzi2007robotic, cha2015perceived, powers2006advisor}. This gap in expectation can significantly reduce human-robot team performance~\cite{groom2007can}.

\begin{figure}[t!]
\centering
\begin{subfigure}[b]{0.90\linewidth}
\centering
 \includegraphics[width=\linewidth]{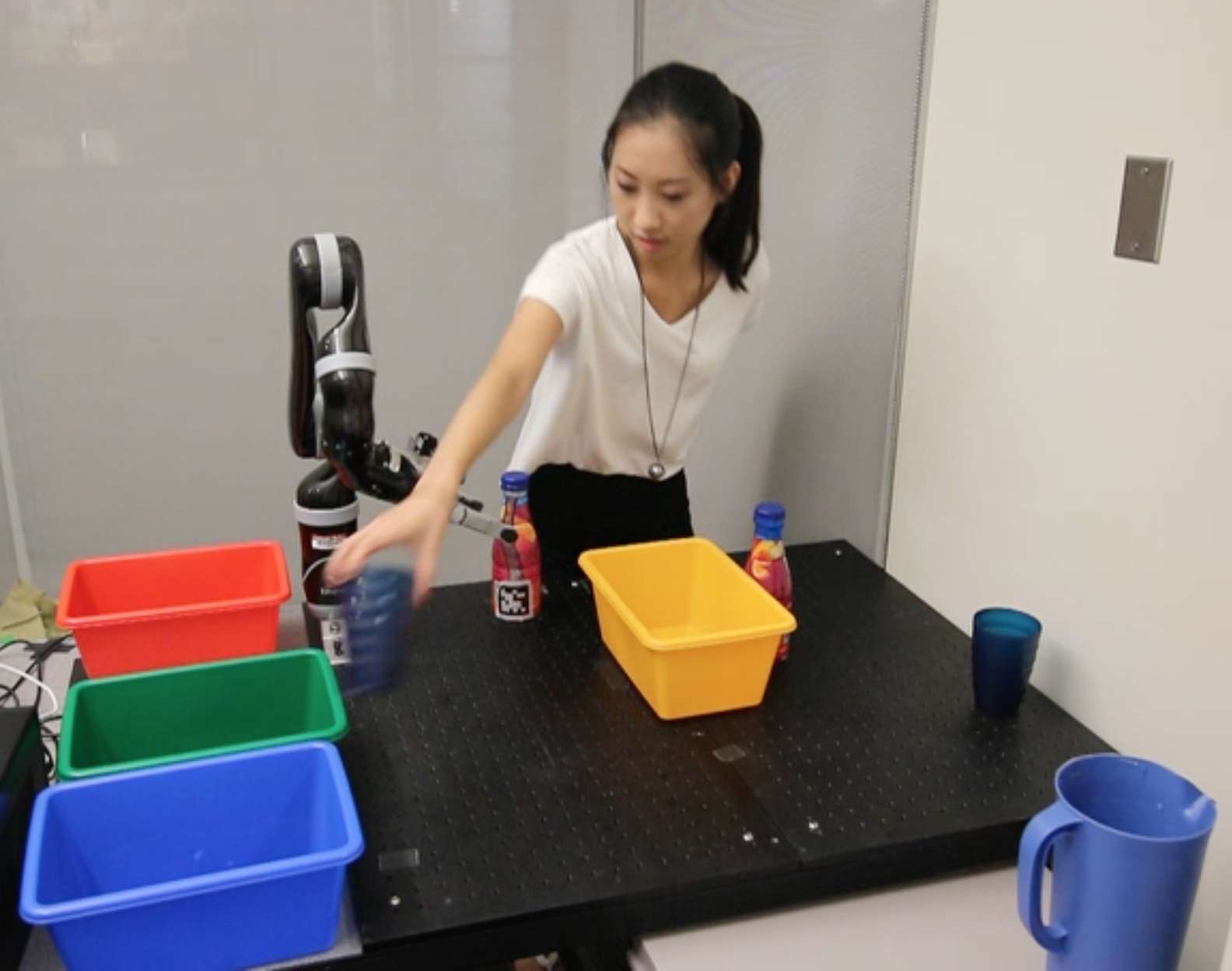}
 \label{fig:table-clearing}
\end{subfigure}
\begin{subfigure}[b]{1.0\linewidth}
\centering
\includegraphics[width=0.447\linewidth]{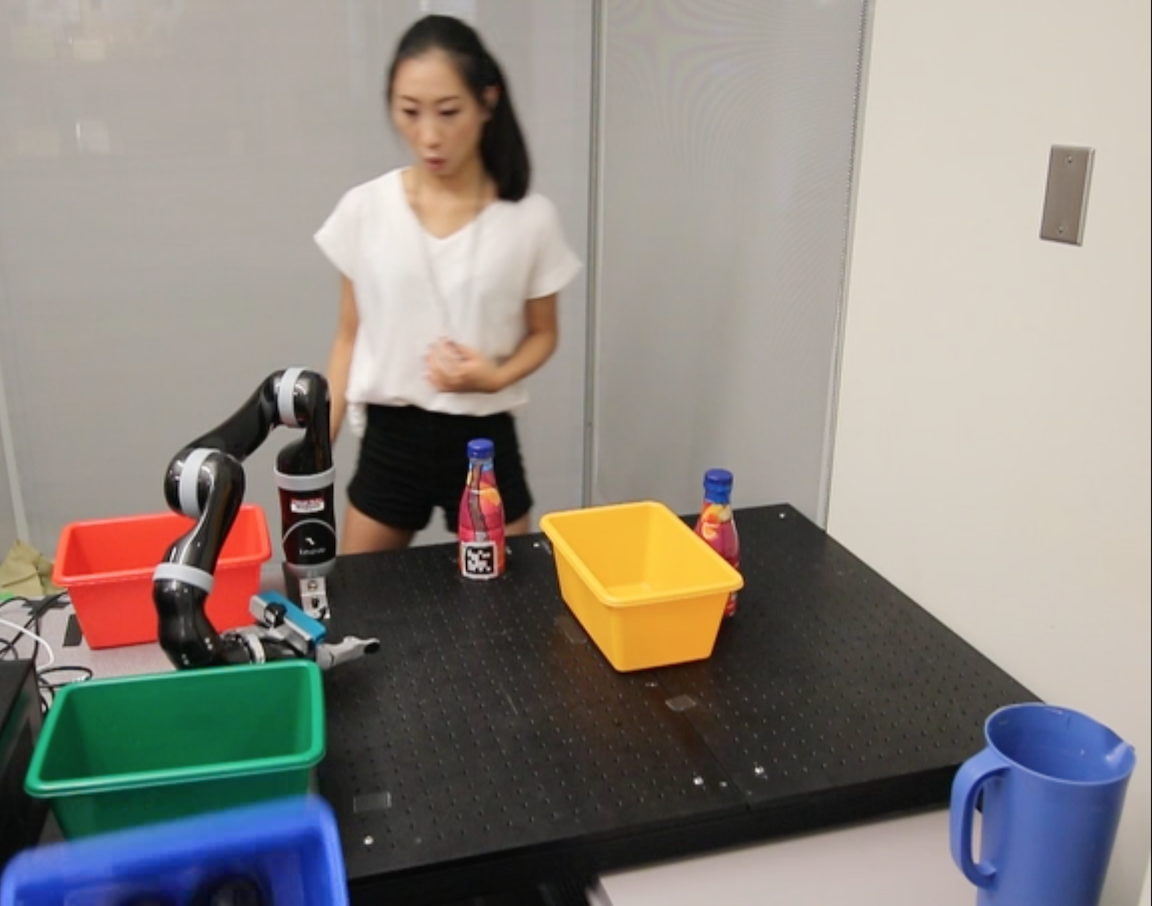}
\includegraphics[width=0.442\linewidth]{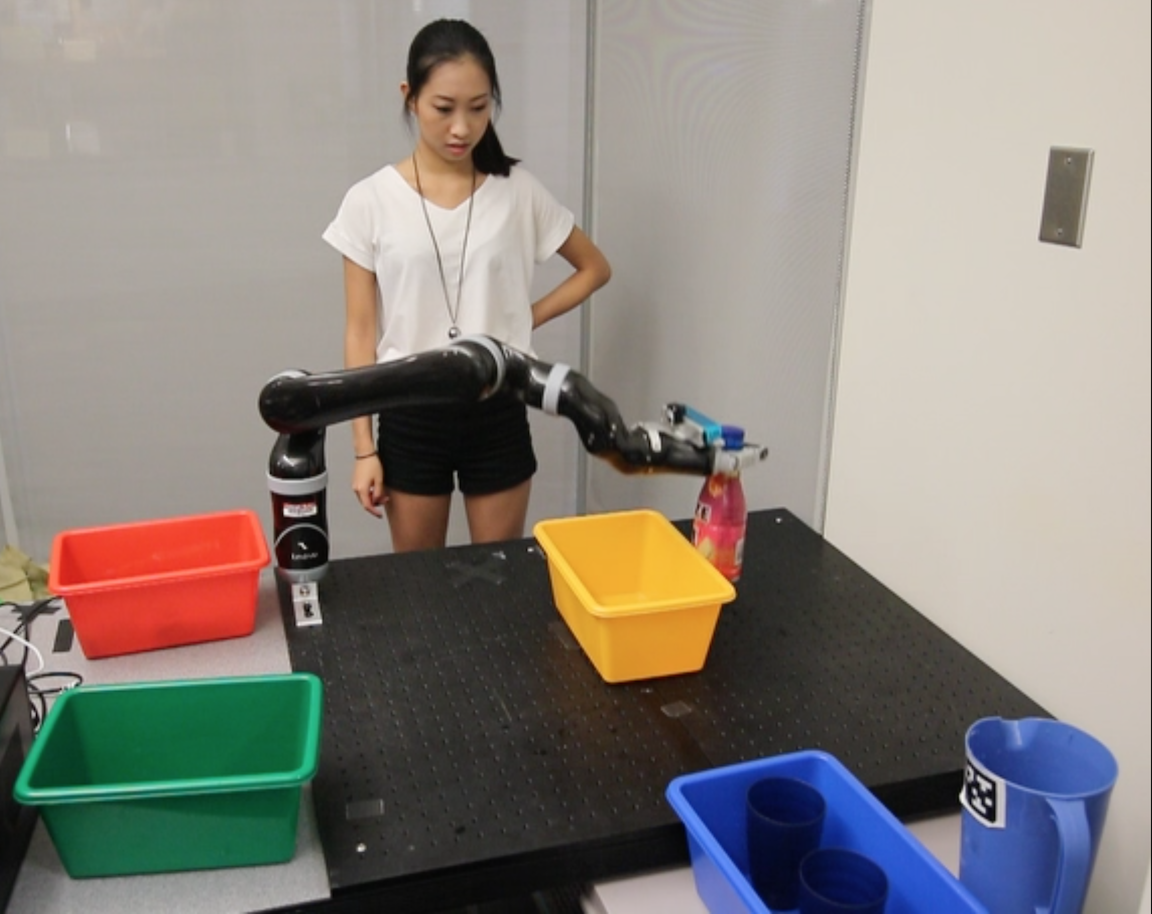}

 \label{fig:robot-fail}
\end{subfigure}
\caption{Top: User performs a repeated table-clearing task with the robot. The robot fails intentionally in the beginning of the task, in order to reveal its capabilities to the human teammate. Bottom-left: The robot drops the blue bin off the table while moving towards the left bottle. Bottom-right: The torques applied exceed their limits when the robot attempts a grasp at an extended configuration, and the robot stops moving.}
\label{fig:front}
\vspace{-0.65cm}
\end{figure}

For example, we consider the table-clearing task illustrated in \cref{fig:front}. The user and the robot are tasked with clearing the table by placing items in the bins. The clearing task is repeated a number of times. We call each repetition a \textit{round}. The user lacks the following information about the robot:
%
\squishlisttwo
\item The robot does not know where the green bin is. If the robot moves, it may collide with the green bin, inadvertently pushing the adjacent blue bin off the table.
\item The robot cannot lift the bottle that is farthest away from its base: the bottle is filled with water and the torques required for a lifting motion exceed the robot's motor torque limits. If the robot attempts to lift the bottle, the robot's control software will abort and the robot will stop moving.
\squishend

\smallskip
\noindent
Nikolaidis et al.~\cite{nikolaidis2016formalizing} proposed a human-robot mutual adaptation formalism, where the robot builds a model of human adaptation to guide the user towards an optimal \textemdash~with respect to some objective performance metric \textemdash~way of completing the task. Every time the human and the robot take an action, the user was modeled as either \textit{completely adopting the robot policy as her\footnote{We use the female pronoun for the human, and the neuter pronoun for the robot.}own} with some probability, or keeping her current policy. This probability was defined as the user's adaptability, which indicated her willingness to adapt to the robot. The formalism allowed the robot to infer the adaptability of its teammate through interaction, and guide the user towards an optimal policy unknown to them in advance. 

While in \cite{nikolaidis2016formalizing}, the human was modeled as completely adopting the robot's optimal policy with some probability, in many collaborative settings human adaptation can be more subtle. We use the table-clearing task described above as an example, and we let the robot attempt to grasp the bottle that is closest to its base, dropping the blue bin in the process. This will likely cause the human teammate to change her actions: in the next round, she will move the green or blue bin out of the robot's way. However, without observing the robot fail in lifting the other bottle, she still has no information about which action to take (i.e. emptying the bottle of water), if the robot attempts to lift the bottle.

This is an example, where \textit{the human may change their actions based on the robot actions, while not completely adopting the robot's optimal policy}.

 In this paper, we propose a game-theoretic model of human partial adaptation to the robot. We assume that the robot knows a ``true'' objective metric of team performance in the form of a reward matrix. We base this assumption on insights from early work on Stackelberg security games, which used domain expert knowledge to specify the reward of the defender/leader (AI agent) and the attacker/follower (human), showing remarkable results~\cite{tambe2011security}.   


We model the human as following a best-response strategy to the robot action, based on their own, possibly distorted, reward function. The human reward function changes over time, as the human observes the outcomes of the robot and her own actions.

 The model allows the robot to \textit{reason over how the human expectations of the robot capabilities will change based on its own actions}. The robot uses this model to compute an optimal policy, which enables it to decide optimally between \textit{revealing information} to the human and \textit{choosing the best action given the information that the human currently has}.

  We prove that, if the robot can observe whether the user has learned at each round, the computation of the optimal policy is simple (\Cref{thm:uniform-row-full-observable,thm:uniform-row-partial-observable}), and can be done in time polynomial in the number of robot actions and the number of rounds (\Cref{thm:efficient-algo}). 

  We show through a human subject experiment in a table-clearing task that the proposed model significantly improves human-robot team performance, compared to policies that assume complete human adaptation to the robot. Additionally, we show through simulations that the proposed model performs well for a variety of randomly generated tasks. This is the first step towards modeling the change of human expectations of the robot capabilities through interaction, and integrating the model into robot decision making in a principled way.

\section{Relevant Work} \label{sec:relevant-work} 
A lot of research in robotics has focused on one-way robot adaptation to the human, where the robot learns a human skill or preference~\cite{Argall2009,AtkesonS97,Abbeel04,Nicolescu03naturalmethods,Chernova2008,AkgunCYT12, Nikolaidis2013}.
Other approaches enable robots to reduce the uncertainty over human intention through information-seeking actions~\cite{Lemon2012, broz2011designing,bandyopadhyay2013intention,macindoe2012pomcop,nikolaidis2015efficient}, through negotiation with the human~\cite{karpas2015robust}, or through decomposition of a complex task into subtasks~\cite{nguyen2011capir}. There has also been work in human adaptation to the robot in social~\cite{goodrich2007human, kanda2004interactive,robins2004effects, Green_makinga}, and physical human-robot interaction~\cite{ikemoto2012physical}, as well as in adaptation between teammates in multi-agent ad hoc team settings~\cite{stone2010ad,stone2010teach}.

Li et al.~\cite{li2015role} suggest that the human-robot collaboration problem in physical human-robot interaction can be modeled as a two-player game. They assume that the human partner exerts a force by optimizing an unknown cost function; the robot's cost-function is then updated based on the gradient of the error between the actual force applied by the human and the force predicted by the robot's cost function, until an equilibrium is achieved. Menell et al.~\cite{hadfield2016cooperative} define a cooperative inverse reinforcement learning (CIRL) problem as a partial information two-player game, where the robot maximizes the unknown human reward in a cooperative setting. They show that solving the game results in active teaching and active learning behaviors. The framework has yet to be evaluated in a human subject experiment. In contrast to both papers, in our work the roles are reversed, since the human learns the robot reward through interaction. In a repeated collaborative task with different actions, human adaptation can be more subtle than in a force exchange scenario. Additionally, the learner (human) does not run an inverse reinforcement learning algorithm. Instead, we model the human as learning with some probability the best-response to the robot action observed. This captures how human actions change over time based on their updated expectations of the robot capabilities, and it enables the robot to decide optimally between communicating the true rewards to the human and maximizing the immediate reward given the current human strategy.

There is also relevant work in the social navigation domain: In the manuscript by Trautman et al.~\cite{trautman2010unfreezing}, human and robot trajectories are jointly planned as the optimum of a reward function that combines goal completion and collision avoidance. Sadigh et al.~\cite{sadigh2016planning} model the interaction of a human driver with an autonomous car as a dynamical system, where the human follows a best-response strategy to the robot actions. By contrast, we focus on a repeated task in a collaborative setting where the human reward function may change over time, as the human observes the outcomes of the robot and her own actions.

We draw upon insights from previous work on a particular class of Stackelberg games~\cite{conitzer2006computing}, the \textit{repeated Stackelberg security games}~\cite{balcan2015commitment}. In this setting, the follower observes the leader's possibly randomized strategy, and chooses a best-response. We extend this model to a human-robot collaboration setting, where the leader is the robot and the follower is the human, and we model human adaptation by having the follower's reward stochastically changing over time. 



\vspace{-2mm}
\section{Formal Model}
Consider a two-player game represented by the player set $N = \{\text{R}, \mathrm{H}\}$, where player $\mathrm{R}$ is the {\tt Robot} and player $\mathrm{H}$ is the {\tt Human}. Each of them has a \emph{finite} set of actions denoted by $A^\mathrm{R} = \{a^\mathrm{R}_1, \ldots, a^\mathrm{R}_m\}$ and $A^\mathrm{H} = \{a^\mathrm{H}_1, \ldots, a^\mathrm{H}_n\}$ respectively. The payoff associated with each pair of actions is uniquely identified by a matrix $R = [r_{i,j}], (i,j) \in [m]\times[n]$, where the entry $r_{i,j}$ denotes the reward\footnote{We will use the terms `reward' and `payoff' interchangeably.} for the action pair $(a^\mathrm{R}_i,a^\mathrm{H}_j)$ chosen by these two players. We denote the reward vector corresponding to row $i$ by $r_i$, i.e., $r_i = (r_{i,1},\ldots,r_{i,n})$. Importantly, the {\em same reward} is experienced together by both players. Therefore this is an {\em identical payoff} game where the goal is to maximize the total reward obtained in $T$ (finite) rounds of playing this repeated game. If the reward matrix was perfectly known to both the agents, they would have played the action pair that gives the maximum reward in each round. 
%
%
%

However, we assume that in the beginning of the game, the robot has perfect information about the reward matrix, whereas the human has possibly incorrect information (captured by a reward matrix $R^H$ which the human {\em believes} to be the true reward matrix). In different rounds of the game, the human probabilistically learns different entries of this matrix and picks action accordingly. We will assume that the human is capable of taking the optimal action given her knowledge of the payoffs, e.g., if a specific row of this matrix is completely known to the human and the robot plays the action corresponding to this row,\footnote{We will refer to this robot action as {\em playing a row}.} the human will pick the action that maximizes the reward in this row.  However, if the entries of a row are yet to be learned by the human, the human picks an action according to $\arg\,\max r^H_i$, where $r^H_i$ is the $i$-th row of $R^H$. 

The only aspect of this game that may change over time is the state of the human, which we denote by $x_t, t \in [T]$. Therefore, the state of the game is simply the state of the human agent. We denote the state space of the game as ${\cal X}$; it will be instantiated below in different models of information dissemination. 


 A policy $\pi = (\pi_1,\ldots,\pi_T)$ is a sequence of robot action functions $\pi_t: {\cal X} \to A^{\mathrm{R}}, \ t \in [T]$. 
 The decision problem of the robot is to find the optimal policy $\pi^* = (\pi^*_1,\ldots,\pi^*_T)$ that maximizes the expected payoff $U_1$ starting from round $1$, defined as follows. Denoting the strategy of the human by $s^{\mathrm{H}} : A^{\mathrm{R}}\times{\cal X} \to A^{\mathrm{H}}$.
%
%
%
\begin{equation}
\label{eqn:decision-problem}
 \begin{aligned}
  U_1(\pi|x_1) &\triangleq \mathbb{E} \left[\left. \sum_{t=1}^T R(\pi_t(x_t),s^{\mathrm{H}}(\pi_t(x_t),x_t)) \right| x_1 \right] \\
  \pi^* &\in \argmax{\pi} \ U_1(\pi|x_1)
 \end{aligned}
\end{equation}
%

\vspace{-2mm}
\section{Approach} \label{sec:approach}
We consider a setting where, in each round, the robot plays first by choosing a row. We model the strategy of the human  $s^{\mathrm{H}} : A^{\mathrm{R}}\times{\cal X} \to A^{\mathrm{H}}$ as maximizing a human reward function $R^{\mathrm{H}}$. In other words, the human \emph{best responds} to the robot action, according to the (possibly erroneous) way she currently perceives the payoffs. The human reward matrix $R^{\mathrm{H}}$ evolves over time, as the human learns the ``true'' reward $R$ through interaction with the robot. We propose a model of human \textit{partial adaptation}, where the human learns with probability $\alpha$ the entries of row $r_i$ that correspond to the robot action $a^\textrm{R}_i$ played, and with probability $(1-\alpha)$ none of the entries. We consider the following models, based on when the human learning occurs, and on whether the robot directly observes if the human has learned. 

\medskip

\noindent$\mathbf{M_1.}$ The human learns the payoffs immediately after the robot plays a row, and before she takes her own action. The robot can infer whether the human has learned the row, by observing the reward after the human has played in the same round. We call this \textit{learning from robot action}, where the robot has \textit{full observability} of the human internal state. This model is studied in Sec.~\ref{subsec:learning-from-robot-action}.

\medskip

\noindent$\mathbf{M_2.}$ The human learns the payoffs associated with a row after she plays in response to the robot's action. The robot can observe whether the human has learned before the start of the next round, for instance by directly asking the human, or by interpreting human facial expressions and head gestures~\cite{el2005real}. We call this model \textit{learning from experience}, where the robot has \textit{full observability} of the human internal state. This model is studied in Sec.~\ref{subsec:learning-from-experience}.

\medskip

\noindent$\mathbf{M_3.}$  Identically to model $\mathbf{M_2}$, the human learns a row after her action in response to the robot action. However, the robot does not immediately observe whether the human has learned, rather infers it through the observation of human actions in subsequent rounds of the game. This is a case of \textit{learning from experience}, \textit{partial observability}.

\medskip

We note that we do not define a model for \textit{learning from robot action}, \textit{partial observability} case, since the robot can always directly observe whether the human has learned, based on the reward resulting from the human action in the same round. 

\begin{figure}[t!]
 \centering
\includegraphics[width=1.0\columnwidth]{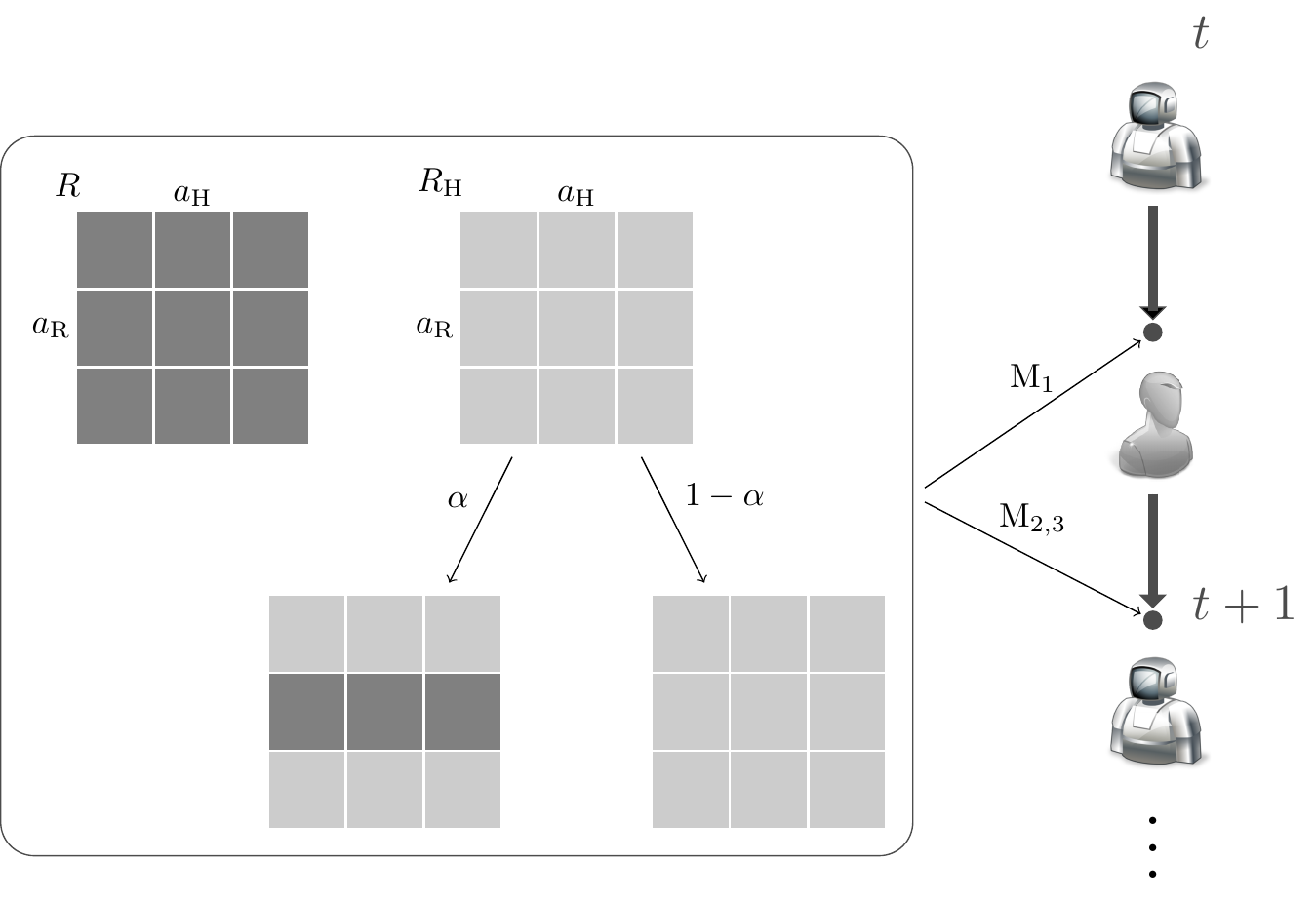}
\caption{Models of human partial adaptation, described in Sec.~\ref{sec:approach}. The human learns with probability $\alpha$ the entries of row $r_i$ that correspond to the robot action $a^\textrm{R}_i$ played, and with probability $1-\alpha$ none of the entries. The learning occurs before her action (\textit{learning from robot action} -- $\mathbf{M_1}$), or after her action (\textit{learning from experience} -- full observability ($\mathbf{M_2}$) or partial observability ($\mathbf{M_3}$)).} 
\label{fig:models}
\vspace{-5mm}
\end{figure}

Figure~\ref{fig:models} shows the different models. In Section~\ref{sec:partial-observability}, we discuss the general case of partial observability (Model $\mathbf{M_3}$) and formulate the problem as a Markov Decision Process~\cite{russel03modern}. Computing the optimal policy in this case is exponential in the number of robot actions $m$. However, when the robot has full observability of the human state (Models $\mathbf{M_1, M_2}$), the optimal policy has a special structure and can be computed in time polynomial in $m$ and $T$ (Section~\ref{sec:full-observability}).

\vspace{-2mm}
\section{Theory: Partial Observability} \label{sec:partial-observability}

In this section we examine the hardest case, where the human learns the payoffs associated with the row after their choice of actions, and the robot cannot directly observe whether the human has learned the payoffs (model $\mathbf{M_3}$). Instead, the robot infers whether the human has learned the row by observing the human response in subsequent rounds of the game. 

While the human state is partially observable, we can exploit the structure of the problem and reduce it to a Markov Decision Process based on the following observation: the probability of the human having learned a row is either 0 when it is played for the first time; $\alpha$ after it is played by the robot and the human responds sub-optimally; and $1$ after the the human has played the actual best-response strategy (according to $R$) for that row (which means she has learned the true rewards in the previous round).

We define a Markov decision process in this setting as a tuple $\{\mathcal{X}, A^{\textrm{R}}, P, R, T\}$, where:
\squishlisttwo
\item $\mathcal{X} \in \{0,\psi,1\}^m $ is a finite set of human states.  A state $x$ is represented by a vector $(x_1, x_2, ... , x_m)$, where $x_i \in \{0,\psi,1\}$ and $i$ is the corresponding row in the matrix. The starting state is $x_{i} = 0$ for each row $i$. $x_{i} = \psi$ indicates that the robot does not know whether human has learned row $i$ or not. In this state, the human plays the best response in that row with probability $\alpha$, or an action defined by the strategy $s^{\mathrm{H}}$ of the human with probability $(1-\alpha)$. If the human plays best-response, then the robot knows that human has learned row $i$, thus the entry for that row is $x_i = 1$.







\item $A^{\textrm{R}}=\{a^\mathrm{R}_1, \ldots, a^\mathrm{R}_m\}$ is a finite set of robot actions.

\item $P : \mathcal{X} \times A^{\textrm{R}} \longrightarrow \Pi(\mathcal{X})$ is the state transition function, indicating the probability of reaching a new state $x'$ from state $x$ and action $a^\textrm{R}_i$. State $x$ transitions to a new state $x'$ with all vector entries identical, apart from the element $x_i$ corresponding to the row played. If the robot plays $i$ for the first time ($x_i = 0$), the corresponding entry in the next state $x'$ deterministically becomes $x'_i = \psi$, since the robot no longer knows whether the human has learned the payoffs for that row. If $x_i = \psi$, the human may have learned that row in the past and play the best-response strategy, leading to a transition to $x'_i = 1$ with probability $\alpha$.  
If the human does not play the best-response strategy, the robot still does not know whether they will have learned the payoffs after the current round, thus $x'_i = \psi$ with probability $(1-\alpha)$. If $x_i = 1$, the corresponding entry in all subsequent states will be $x'_i = 1$, i.e., if the human learns a row, we assume that she remembers the row in the future.

\item $R :  A^\mathrm{R} \times A^\mathrm{H} \longrightarrow \mathbb{R}$ is the reward function, giving the immediate reward gained by performing a human and robot action. 
Note that if action $i$ is played and the state has $x_i=\psi$, the reward will be based on the best response in row $i$ of $R$ with probability $\alpha$, and on row $i$ of $R^{\mathrm{H}}$ with probability $(1-\alpha)$ --- we consider the \emph{expected} reward.

\item $T$ is the number of rounds.
\squishend

\smallskip
\noindent
The robot's decision problem is to find the optimal policy $\pi^* = (\pi^*_1,\ldots,\pi^*_T)$ 
to maximize the expected payoff $U_1$ as defined in Eq.~\ref{eqn:decision-problem}. 

We observe that in the current formalism, the size of the state-space is $|\mathcal{X}| = 3^m$, where $m$ is the number of robot actions. Therefore, the computation of the optimal policy requires time exponential in $m$. In Section~\ref{sec:full-observability}, we show that for the case where the robot can observe whether the human has learned the payoffs, the optimal policy can be computed in time polynomial in $m$ and $T$.
%
%

\section{Theory: Full Observability} \label{sec:full-observability}

In this section, we assume that the robot can observe whether the human has learned the payoffs.
We instantiate state $x_t$ as a vector $(x_{t,1},x_{t,2},\ldots,x_{t,m})$, where each $x_{t,i}$ is now a binary variable in $\{0,1\}$ denoting the robot's knowledge in round $t$ of whether row $i$ is learned by the human. In contrast to Sec.~\ref{sec:partial-observability}, there is no uncertainty about whether the human has learned or not (therefore no $\psi$ state). 

\subsection{Learning from Robot Action} \label{subsec:learning-from-robot-action}
This is the scenario where the human might learn the payoffs immediately after the robot plays a row, and before she takes her own action (Model $\mathbf{M_1}$ in Sec.~\ref{sec:approach}). Clearly, the robot can figure out if the human learned the row by observing the reward for that round. Our algorithmic results in this model strongly rely on the following lemma. 
%
\begin{lemma}
 \label{thm:uniform-row-full-observable}
 In model $\mathbf{M_1}$, if, under the optimal policy $\pi^*$, there exists $\tau \in \{2,\ldots,T\}$ and $i \in [m]$ such that $x_{\tau,i} = 1$ and $\max r_i \geqslant \max r_j$ for all $j$ such that $x_{\tau,j} = 1$, then $\pi^*_t(x_t) = a^\mathrm{R}_i$ for all $\tau \leqslant t \leqslant T$ and for all $x_t = x_\tau$.
\end{lemma}
This lemma says that the optimal policy for the robot is to pick the action $a^\mathrm{R}_i$ when $i$ is the row that yields the maximum reward among the rows already learned by the human. As we will show in detail later, this directly leads to a computationally efficient algorithm, via the following insight: \textit{if the robot plays a row and this row is successfully revealed to the human, the optimal policy for the robot is to keep playing that row until the end of the game.}

The main idea behind the proof below is: if at round $t-1$ the optimal
policy plays row 2, and that row is revealed, then it will not explore
the unrevealed (higher rewarding) row 1 afterwards. The reason is that if the optimal policy
chose to explore row 1 at some time in the future --- which is a
contradiction to the lemma ---  then playing row 1 at
round $t-1$ would have been optimal, therefore an optimal policy would not have
played row 2 at round $t-1$.


\begin{proof}[of Lemma~\ref{thm:uniform-row-full-observable}]
 Assume for contradiction that the lemma does not hold, and let $t$ be the \emph{last} round in which the optimal policy violates the lemma, i.e., the last round in which there are $i,j\in [m]$ such that $x_{t,i}=0$ and $x_{t,j}=1$, but the optimal policy plays row $i$. Without loss of generality assume that these $i$ and $j$ are rows $1$ and $2$, respectively. For all rounds from $t+1$ to $T$, it holds (by the choice of $t$) that if row $i$ is revealed to the human, the optimal policy will continue playing $a^\mathrm{R}_i$ (if there are multiple such rows, it plays the one with highest reward).

 Let the maximum rewards corresponding to rows $1$ and $2$ be $R_1$ and $R_2$, respectively, i.e., $R_k = \max r_k$. We assume w.l.o.g.~that row $2$ has the highest maximum reward among all revealed rows. We can also assume that $R_1 > R_2$, since a policy that moves away from a row that is simultaneously known and more rewarding is clearly suboptimal.
 
 If a row is not learned, the reward associated with actions $a^\mathrm{R}_1$ and $a^\mathrm{R}_2$ are $C_1$ and $C_2$, where $C_k = r_k [\arg\!\max r^H_k]$. Clearly, $C_1 \leqslant R_1$ and $C_2 \leqslant R_2$.
%
 Since the optimal policy chose $a^\mathrm{R}_1$ in round $t$ over $a^\mathrm{R}_2$, the expected payoff of choosing $a^\mathrm{R}_1$ in round $t$ must be larger than that of $a^\mathrm{R}_2$, i.e.,
{\small
\begin{equation*}
\begin{split}
  &\alpha (R_1 + U_{t+1}(\pi^* | (1,1, \ldots))) + (1-\alpha) \cdot  
   (C_1 + U_{t+1}(\pi^* | (0,1, \ldots)))\\
&\quad > R_2 + U_{t+1}(\pi^* | (0,1, \ldots)),
\end{split}
\end{equation*} }
where the first term on the LHS shows the expected payoff if row $1$ is learned in round $t$, and the second term shows the payoff when it is not. It follows that
{\small
\begin{equation}
\begin{split}
&\alpha (R_1 + R_1 \cdot (T-t-1)) + (1-\alpha) \cdot 
   (C_1 + R_2 \cdot (T-t-1))\\
&\quad  > R_2 + R_2 \cdot (T-t-1).
\end{split}
   \label{eqn:implication}
\end{equation} }
 The implication holds because from round $t+1$, we assume (by the choice of $t$) that the optimal policy continues playing the best action among the revealed rows. We make the above inequality into an equality by adding a slack variable $\epsilon > 0$ as follows.
 \begin{align}
  \label{eqn:slack}
 &\alpha R_1 \cdot (T-t) + (1-\alpha)(C_1 + R_2 \cdot (T-t-1)) \nonumber \\ 
  &\quad = R_2 + R_2 \cdot (T-t-1) + \epsilon.
 \end{align}
 Denote the LHS of the above equality as $\rho_1$. Note that this is the assumed optimal value of the objective function at round $t$ when the state $x_t$ is $(0,1,\ldots)$, i.e., $U_t(\pi^* | (0,1,\ldots)) = \rho_1$. Rearranging the expressions above, we get,
 \begin{equation}
 \label{eqn:lemma1-expr1}
  \alpha R_1 \cdot (T-t) + (1-\alpha)C_1 = R_2 + \alpha R_2 \cdot (T-t-1) + \epsilon. 
 \end{equation}
 We claim that if the optimal policy chooses the action $a^\mathrm{R}_1$ at round $t$, then the expected payoff in round $t-1$ from choosing the action $a^\mathrm{R}_1$ would have been larger than that of the action $a^\mathrm{R}_2$. If our claim is true, then the current policy, which chose $a^\mathrm{R}_2$ at $t-1$, cannot be optimal, and we reach a contradiction. To analyze the decision problem in round $t-1$, we need to consider two possible states of the game in this round.
 
\medskip

 \noindent {\em Case 1: $x_{t-1} = (0,0,\ldots)$}. In this state, playing $a^\mathrm{R}_1$ gives an expected payoff of
 \begin{equation}
 \label{eqn:lemma1-expr2}
 \begin{aligned}
  \lefteqn{\alpha (R_1 + U_t(\pi^* | (1,0,\ldots))) + (1-\alpha)(C_1 + U_t(\pi^* | (0,0,\ldots))) } \\
  &\quad \geqslant \alpha (R_1 + R_1 (T-t)) + (1-\alpha)(C_1 + U_t(\pi^* | (0,0,\ldots))).
  \end{aligned}
 \end{equation}

 The inequality holds because in state $(1,0,\ldots)$, playing $a^\mathrm{R}_1$ yields at least $R_1$ in every subsequent round.
 Playing $a^\mathrm{R}_2$ in round $t-1$ yields,
 \begin{equation}
  \label{eqn:lemma1-expr3}
  \alpha (R_2 + \rho_1) + (1-\alpha)(C_2 + U_t(\pi^* | (0,0,\ldots))).
 \end{equation}
 This expression is similar to the RHS of \Cref{eqn:lemma1-expr2}, except that the expected payoff at $x_t = (0,1,\ldots)$ is assumed to be $\rho_1$. 
 We claim that the expression on the RHS of \cref{eqn:lemma1-expr2} is larger than the expression in \cref{eqn:lemma1-expr3}, for which we need to show that
 \begin{alignat*}{2}
  &\alpha (R_1 + R_1 (T-t)) + (1-\alpha)C_1 \\
  &\quad > \alpha (R_2 + \rho_1) + (1-\alpha)C_2 \\
  \Longleftrightarrow\ & \alpha R_1 + R_2 + \alpha R_2 \cdot (T-t-1) + \epsilon \\
  &\quad > \alpha (R_2 + R_2 + R_2 \cdot (T-t-1) + \epsilon) + (1-\alpha)C_2 \\
  \Longleftrightarrow\ & \alpha R_1 + R_2 + \epsilon > \alpha R_2 + \alpha R_2 + (1-\alpha)C_2 + \alpha \epsilon.
 \end{alignat*}
 In the first equivalence, we substitute the expression from \cref{eqn:lemma1-expr1} on the LHS and the expression of $\rho_1$ from \cref{eqn:slack} on the RHS. The second equivalence holds by canceling out one term. We see that the final inequality is true since $R_2 \geqslant C_2, R_1 > R_2$, and $0 < \alpha < 1$.\footnote{If $\alpha = 1$, playing the row $\arg \max{R_i}$ is optimal and the lemma holds trivially. For $\alpha = 0$, the lemma is vacuously true. So, we assume $0 < \alpha < 1$ w.l.o.g.}

\medskip
 
 \noindent {\em Case 2: $x_{t-1} = (0,1,\ldots)$}, in this state playing the action $a^\mathrm{R}_1$ gives an expected payoff of at least
 \begin{align}
  &\alpha (R_1 + R_1 \cdot (T-t)) + (1-\alpha)(C_1 + U_t(\pi^* | (0,1, \ldots))) \nonumber \\
  &\quad = \alpha (R_1 + R_1 \cdot (T-t)) + (1-\alpha)(C_1 + \rho_1). \label{eqn:lemma1-expr4}
 \end{align}
 This is similar to the RHS of \cref{eqn:lemma1-expr2} except that now we can replace $U_t(\pi^* | (0,1, \ldots))$ with $\rho_1$. On the other hand, the expected payoff of the action $a^\mathrm{R}_2$ in round $t-1$ is given by $R_2 + \rho_1$ --- because at state $(0,1, \ldots)$ in round $t-1$, action $a^\mathrm{R}_2$ gives $R_2$ deterministically, since the human knows row 2. The state remains the same even after reaching round $t$. The expected payoff at this round for this state is assumed to be $\rho_1$.
 Now to show that the expression in \cref{eqn:lemma1-expr4} is larger than $R_2 + \rho_1$, we need to show that
 \begin{align*}
  &\alpha (R_1 + R_1 \cdot (T-t)) + (1-\alpha)(C_1 + \rho_1) > R_2 + \rho_1 \\
  \Longleftrightarrow\ & \alpha R_1 + \alpha  R_1 \cdot (T-t) + (1-\alpha)C_1 > R_2 + \alpha \rho_1 \\
  \Longleftrightarrow\ & \alpha R_1 + R_2 + \alpha R_2 \cdot (T-t-1) + \epsilon \\
  &\quad > R_2 + \alpha R_2 + \alpha R_2 \cdot (T-t-1) + \alpha \epsilon \\
  \Longleftrightarrow\ & \alpha R_1 + \epsilon > \alpha R_2 + \alpha \epsilon
 \end{align*}
 The first equivalence comes from reorganizing the inequality. The second equivalence is obtained through substitution using \cref{eqn:slack,eqn:lemma1-expr1}. The third equivalence follows by canceling out two terms. The last inequality is true since $R_1 > R_2$ and $0 < \alpha < 1$.

 
 To summarize, we have reached a contradiction in both cases, which are exhaustive. This proves the lemma.
\end{proof}

\subsection{Learning from Experience} \label{subsec:learning-from-experience}

Recall that in model $\mathbf{M_2}$, the human learns with probability $\alpha$ all payoffs associated with a row {\em after} she plays her action in response to the robot playing an unrevealed row. She does not learn with probability $1-\alpha$. This model is the same as model $\mathbf{M_3}$ of Sec.~\ref{sec:partial-observability}, with an additional assumption: before the robot takes its next action, it can observe the current state.  

We show that in this setting too, the optimal policy has a special structure similar to that under model $\mathbf{M_1}$ (Sec.~\ref{subsec:learning-from-robot-action}), which can be computed in time polynomial in $m$ and $T$.
%

\begin{lemma}
 \label{thm:uniform-row-partial-observable}
 In model $\mathbf{M_2}$, if, under the optimal policy $\pi^*$, there are $\tau \in \{2,\ldots,T\}$ and $i \in [m]$ such that $x_{\tau,i} = 1$ and $\max r_i \geqslant \max r_j$ for all $j$ such that $x_{\tau,j} = 1$, then $\pi^*_t(x_t) = a^\mathrm{R}_i$ for all $\tau \leqslant t \leqslant T$ and for all $x_t = x_\tau$.
\end{lemma}

The proof is similar to the proof of \Cref{thm:uniform-row-full-observable}. However, the expected payoffs and the corresponding inequalities are different. Therefore, we provide a proof sketch that identifies the differences from the previous proof.

\begin{proof}[of Lemma~\ref{thm:uniform-row-partial-observable} (sketch)]
 As before, the idea of the proof is to show that if the optimal policy changes its action from playing the revealed row that yields maximum reward, $a^\mathrm{R}_2$, to playing an unrevealed row of higher maximum reward, $a^\mathrm{R}_1$, for the last time in round $t$, then it must have done so in its previous round, leading to a contradiction. In model $\mathbf{M_2}$, the human does not observe the payoffs of the row played by the robot before she plays her own action. Therefore, we can assume w.l.o.g.~that when an unrevealed row is played, its reward is no larger than the maximum reward of that row, e.g., $C_1 \leqslant R_1$ if row $1$ is played. Hence, if the optimal policy changes its action from $a^\mathrm{R}_2$ to $a^\mathrm{R}_1$ in round $t$ when $x_t = (0,1,\ldots)$, the inequality equivalent to \cref{eqn:implication} must be
 \begin{align}
  &C_1 + \alpha R_1 \cdot (T-t-1) + (1-\alpha) R_2 \cdot (T-t-1) \nonumber \\
  &\quad > R_2 + R_2 \cdot (T-t-1). \label{eqn:implication-2}
 \end{align}
 After adding the slack variable, we get,
 \begin{align*}
  \rho_1 &\triangleq C_1 + \alpha R_1 \cdot (T-t-1) + (1-\alpha) R_2 \cdot (T-t-1) \nonumber \\
   & = R_2 + R_2 \cdot (T-t-1)+ \epsilon \label{eqn:rho} \\
   \implies& C_1 + \alpha R_1 \cdot (T-t-1) = R_2 + \alpha R_2 \cdot (T-t-1)+ \epsilon.
 \end{align*}
 In {\em Case 1}, the expected payoff of playing $a^\mathrm{R}_1$ is at least:
 $C_1 + \alpha R_1 \cdot (T-t) + (1-\alpha) U_t(\pi^* | (0,0,\ldots))$.
 The expected payoff of playing $a^\mathrm{R}_2$ is:
 $C_2 + \alpha \rho_1 + (1-\alpha) U_t(\pi^* | (0,0,\ldots))$.
 We show that the first expression is larger than the second, i.e.,
 \begin{align*}
  &C_1 + \alpha R_1 \cdot (T-t) > C_2 + \alpha \rho_1 \\
  \Longleftrightarrow\ & \alpha R_1 + R_2 + \alpha R_2 \cdot (T-t-1) + \epsilon \\
  &\quad > C_2 + \alpha R_2 + \alpha R_2 \cdot (T-t-1) + \alpha \epsilon \\
  \Longleftrightarrow\ & \alpha R_1 + R_2 + \epsilon > C_2 + \alpha R_2 + \alpha \epsilon.
 \end{align*}
 The final inequality holds since $R_1 > R_2 \geqslant C_2$ and $0 < \alpha < 1$.
 
 Similarly for {\em Case 2}, the expected payoff of playing $a^\mathrm{R}_1$ is at least:
 \begin{align*}
  &C_1 + \alpha R_1 \cdot (T-t) + (1-\alpha) U_t(\pi^* | (0,1,\ldots)) \\
  &\quad \geqslant C_1 + \alpha R_1 \cdot (T-t) + (1-\alpha) R_2 \cdot (T-t).
 \end{align*}
 On the other hand, the expected payoff of playing $a^\mathrm{R}_2$ is $R_2 + \rho_1$. We again show that the RHS of the first expression is larger than the second, i.e.,
 \begin{align*}
   &C_1 + \alpha R_1 \cdot (T-t) + (1-\alpha) R_2 \cdot (T-t) > R_2 + \rho_1 \\
  \Longleftrightarrow\ & C_1 + \alpha R_1 \cdot (T-t-1) + \alpha R_1 + (1-\alpha) R_2 \cdot (T-t-1) \\
  &\quad + (1-\alpha) R_2 > R_2 + R_2 + R_2 \cdot (T-t-1) + \epsilon\\
  \Longleftrightarrow\ & R_2 + R_2 \cdot (T-t-1)+ \epsilon + \alpha R_1 + (1-\alpha) R_2 \\
  &\qquad > R_2 + R_2 + R_2 \cdot (T-t-1)+ \epsilon \\
  \Longleftrightarrow\ & \alpha R_1  > \alpha R_2,
 \end{align*}
 which holds since $R_1  > R_2$ and $0 < \alpha < 1$. 
\end{proof}

\begin{algorithm}[ht]
  \caption{Optimal Policy: Full Observability}
  \label{algo:opt-policy}
 \begin{algorithmic}
 \State {\bf Input:} matrix $R$, time horizon $T$, parameter $\alpha$
 \State {\bf Output:} optimal action $a^*_t$ in each round $t$
 \State $U_t(x_t), a^*_t(x_t) = $ {\tt OptPolicy}($x_t,t$)  
  \Procedure{\tt OptPolicy}{$x_t, t$}
  \If{$t>T$}
    \State \textbf{return} {\tt (0, None)} 
  \Else
    \If{$x_t$ has at least one $1$}
      \State find a row $k^*$ s.t. $k^* \in \argmax{k:x_{t,k}=1}\ \max \ r_{k}$
      \State \textbf{return} {\tt ($\max r_{k^*} \times (T-t), k^*$)}
    \Else
      \State find a row
	  \State 
	  $i^* \in \argmax{k \in [m]} \ [\alpha (R_k + U_{t+1}(\mathbf{e}_k)) + (1-\alpha) (C_k$
	  \State $~~~~~~~~~~~~~~~~~~+U_{t+1}(\mathbf{0}))]$
	  \State and its value $u_{i^*}$ ({\em for model $\mathbf{M_1}$})
      \State {\bf OR}
      \State find a row 
	  \State 
	  $i^*\in \argmax{k \in [m]} \ [C_k + \alpha U_{t+1}(\mathbf{e}_k) + (1-\alpha) U_{t+1}(\mathbf{0})]$
      \State and its value $u_{i^*}$ ({\em for model $\mathbf{M_2}$})
      \State \textbf{return} {\tt ($u_{i^*}, i^*$)}
    \EndIf
  \EndIf
  \EndProcedure
 \end{algorithmic}
\end{algorithm}

\subsection{Design of an efficient algorithm}

As advertised, using \Cref{thm:uniform-row-full-observable,thm:uniform-row-partial-observable}, we can easily prove the following theorem. 

\begin{theorem}
\label{thm:efficient-algo}
In models $\mathbf{M_1}$ and $\mathbf{M_2}$, an optimal policy can be computed in polynomial time.
\end{theorem}

Indeed, the algorithm is specified as \Cref{algo:opt-policy}.
Here $\mathbf{e}_k$ denotes the $m$-dimensional standard unit vector in direction $k$.
This algorithm runs in time polynomial in $m$ and $T$ since the inner {\tt else} condition does not branch into two independent computations.
This is because when at least one coordinate of $x_t$ is $1$, the inner {\tt if} condition is met and the expected payoff in that case is computed without recursion. 
Therefore, in every round the number of computations is $O(m)$, and the algorithm has complexity $O(mT)$. 

\section{From Theory to Users}
We conduct a human subject experiment to evaluate the proposed model in a table-clearing task (Fig.~\ref{fig:front}). We focus on the case where the human \textit{learns from experience} (Models $\mathbf{M_2,M_3}$). We are interested in showing that the policies computed using the partial adaptation model will result in better performance than policies that model the human as completely adapting to the robot, which is an assumption used in previous work on human-robot mutual adaptation~\cite{nikolaidis2016formalizing}.



\begin{figure}[t!]
\centering
\hspace{-0.3cm}
\begin{subfigure}[l]{0.43\linewidth}
\includegraphics[width=1.1\linewidth]{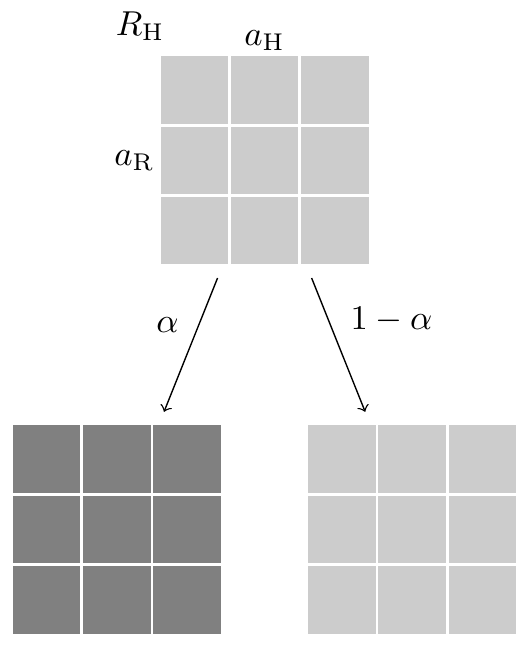}
\caption{}
\label{fig:revealing-simple}
\end{subfigure}
\hspace{0.8cm}
\begin{subfigure}[l]{0.43\linewidth}
\includegraphics[width=1.1\linewidth]{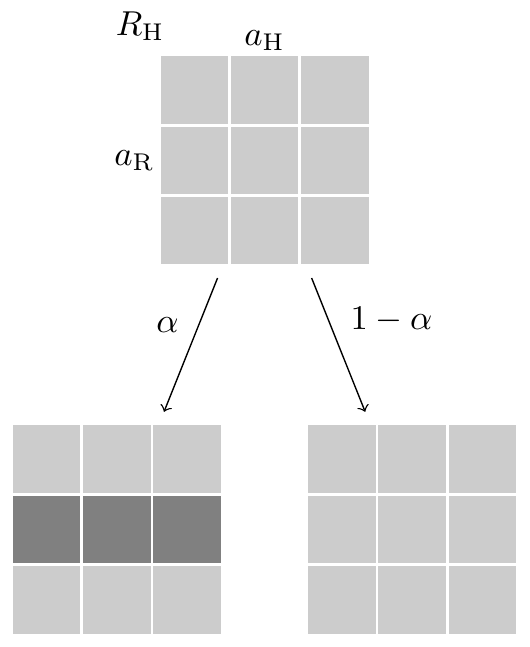}
\caption{}
\label{fig:revealing-medium}
\end{subfigure}
\label{fig:revealing-simple-medium}
\caption{The robot reward matrix $R$ is in dark shade and the human reward matrix $R^\mathrm{H}$ in light shade. (a) The robot reveals its whole reward matrix with probability $\alpha$. (b) The robot reveals the row played (in this example row 2) with probability $\alpha$.}
\end{figure}

\subsection{Manipulated Variables}
\noindent\textbf{Observability.} We used two settings \textemdash~ one where the robot does not directly observe whether the human has learned (Sec.~\ref{sec:partial-observability}), and one where the robot observes directly whether the human has learned (Sec.~\ref{subsec:learning-from-experience}).  \\
\noindent\textbf{Adaptation.} We compared the proposed partial adaptation model with a model of complete adaptation, where the robot models the human as learning all rows of the payoff matrix with probability $\alpha$ after a row is played, instead of learning only the row played (Fig.~\ref{fig:revealing-simple}). 



\subsection{Hypothesis}
\textit{We hypothesize that the robot policies that model the human as partially adapting to the robot will perform better than the policies that assume complete adaptation of the human to the robot.}

\subsection{Experiment Setting}
\noindent\textbf{Table-clearing task.}~
We test the hypothesis in the table-clearing task of Fig.~\ref{fig:front}, where a human and a robot collaborate to clear the table from objects. In this task, the human can take any of the following actions: \{pick up any of the blue cups and place them on the blue bin, change the location of any of the bins, empty any of the bottles of water\}. The robot can either remain idle or pick up any of the bottles from the table and move them to the red bin. The goal is to maximize the number of objects placed in the bins.

The human does not have in advance the following information about the robot: (1) the robot does not know the location of the green bin. Therefore, when the robot attempts to grab one of the bottles, it may push the green bin, dropping the blue bin off the table. (2) The robot will fail if it picks up the bottle that is farthest away from it, if that bottle has water in it. This is because of its motor torque limits. 

\noindent\textbf{Model parameters. }
This information is represented in the form of a payoff matrix $R$. The entries correspond to the number of objects in the bins after each human and robot action. Table~\ref{table:table-clearing-payoff} shows part of $R$; it includes only the subset of human actions that affect the outcome. For instance, if the robot starts moving towards the bottle that is closest to it (action `Pick up closest') and the human does not move the green or blue bin out of the way, the robot will drop the blue bin off the table, together with any blue cups that the human has placed. Therefore, at the end of the task only the bottle will be cleared from the table, resulting in a reward of 1. If the robot attempts to pick up both bottles (action ``pick up both'') and the human does not empty the bottle of water before the robot grasps it, the robot will fail, resulting in a reward of 0. If the human has emptied the bottle and moved the blue bin (action ``Clear cups \& move bin \& empty bottle''), the robot will successfully clear both bottles without dropping the bin, resulting in a reward of 4 (2 bottles in the red bin and 2 cups in the blue bin).


In the beginning of the task, we assume that the human response to all robot actions will be ``Clear cups''; since the human has not observed the robot dropping the bin or failing to pick up the  bottle, she has no reason to move the bin or empty the bottle of water. We also assume that she does not learn any payoffs if the robot remains idle (``Noop'' action). We set the probability of learning $\alpha = 0.9$, since we expected most participants to learn the best-response to the robot actions after observing the outcome of their actions.

\noindent\textbf{Procedure.}
The experimenter first explained the task to the participants and informed them about the actions that they could take, as well as about the robot actions. Participants were told that the goal was to maximize the number of objects placed in the bins at each round. They performed the task three times ($T=3$). In the full observability setting, the experimenter asked the participants after each round, what would their action be if the robot did the same action in the next round. The experimenter then inputted their response (learned or not learned) into the program that executed the policy. When the robot failed to pick up the bottle, the experimenter informed them that the robot had failed. Participants were told that the error message displayed in the terminal was: ``The torque of the robot motors exceeded their limits.'' This is the generic output of our ROS-based hardware interface, when the measured torques exceed the manufacturer limits. We added a short, general explanation about how torque is related to distance and applied force. At the end, participants answered open-ended questions about their experience in the form of a video-taped interview. \\

\newcommand{\specialcell}[2][c]{%
  \begin{tabular}[#1]{@{}c@{}}#2\end{tabular}}

\newcolumntype{C}[1]{>{\centering\let\newline\\\arraybackslash\hspace{0pt}}m{#1}}
\newcolumntype{R}{>{\raggedleft\arraybackslash}p{2cm}}
\newcolumntype{L}{>{\raggedright\arraybackslash}p{1cm}}

\begin{table}[h!]
 \begin{tabular}{L R R R }
   & Clear cups & Clear cups \& move bin & Clear cups \& move bin \&  empty bottle    \vspace{0.5em}
\\ 
  \specialcell{Noop\\} & $2$ & $2$ & $2$ \\ 
  \specialcell{Pick up closest} & $1$ & $3$ & $3$ \\ 
  \specialcell{Pick up both} & $0$ & $0$ & $4$ \\ 
 \end{tabular}
 \caption{Part of payoff matrix $R$ for table-clearing task. The table includes only the subset of human actions that affect performance.}
\label{table:table-clearing-payoff}
\end{table}

\subsection{Subject Allocation}
 We recruited 60 participants from a university campus. We chose a between-subjects design in order to avoid biasing users towards policies from previous conditions. 

\section{Results and Discussion}
\begin{figure*}
 \centering
  \includegraphics[width=1.0\linewidth]{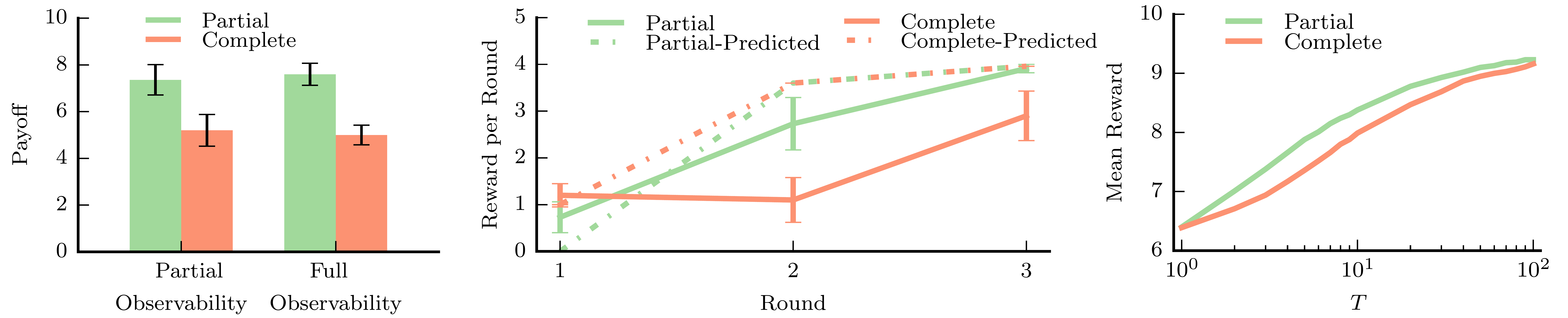}
 \caption{Left: Accumulated reward over 3 trials of the table-clearing task for all four conditions. Center: Predicted and actual reward by the partial and complete adaptation policies in the partial observability setting. Right: Mean reward over time horizon $T$ for simulated runs of the complete and partial adaptation policies in the partial observability setting. The gain in performance from the partial adaptation model decreases for large values of $T$. The x-axis is in logarithmic scale.}
  \label{fig:performance}

 \end{figure*}

\noindent\textbf{Analysis.}
We evaluate team performance by the accumulated reward over the three rounds of the task (Fig.~\ref{fig:performance}-left). We observe that the mean reward in the partial adaptation policy was 42\% higher than that of the complete adaptation policy in the partial observability setting, and 52\% higher than that of the complete adaptation policy in the full observability setting. A factorial ANOVA showed no significant interaction effects between the observability and adaptation factors. The test showed a statistically significant main effect of adaptation ($F(1,56)=18.58, p < 0.001$), and no significant main effect of observability. These results support our hypothesis.

The difference in performance occurred because in the complete adaptation model the robot erroneously assumed that the human had learned the best-response to the ``Pick up both'' action, after the robot played the row ``Pick up closest''. In this section, we examine the partial and complete adaptation policies in the \textit{partial-observability} setting. The interpretation of the robot actions in the \textit{full-observability} setting is similar, and we omit it because of space limitations. The robot chooses the action ``Pick up both'' for round $T=1$ (as well as for $T=2,3$) in the partial adaptation condition\footnote{Unless specified otherwise, for the rest of this section we refer to the partial observability level of the observability factor.}, since the loss of receiving zero reward at $T=1$ is outweighed by the rewards of 4 in subsequent rounds, if the human learns the best-response to that action, which occurs with high probability  ($\alpha = 0.9$). On the other hand, the robot in the complete adaptation condition takes the action ``Pick up closest'' at $T=1$ and ``Pick up both'' at $T=2$ and $T=3$. This is because the model assumes that the human will learn the best-response for all robot actions if the robot plays either ``Pick up closest'' or ``Pick up both'', and the predicted reward of 1 for the action ``Pick up closest'' is higher than the reward of 0 for ``Pick up both'' at $T=1$.

Fig.~\ref{fig:performance}~(center) shows the expected immediate reward predicted by the partial and complete adaptation model for each round in the partial observability setting, and the actual reward that participants received. We see that the immediate reward  in the complete adaptation condition at $T=2$ was significantly lower than the predicted one. The reason is that six participants out of 10 in that condition did not infer at $T=1$ that the robot was unable to pick up the second bottle and did not empty the bottle at $T=2$, which was the best-response action. From the four participants that emptied the bottle, two of them justified their action by stating that ``there was enough time to empty the bottle'' before the robot would grab it. The same justification was given by three participants out of eleven in the partial adaptation condition, who emptied the bottle at $T=1$ without knowing that this was required for the robot to be able to pick it up. This caused the actual reward to be higher than its predicted value of 0. Additionally, the actual reward at $T=2$ was lower than the predicted value. We attribute this to the fact that 73\% of participants learned the best-response for the robot action (emptying the bottle that was farthest away) in that round, whereas the predicted value assumed a probability of learning $\alpha = 0.9$. In $T=3$, the actual reward matched the prediction closely, since all participants eventually learned that they should empty the bottle.

\noindent\textbf{Generalizability of the results.}
The results discussed above are compelling in that they arise from an actual human-subject experiment, but they are limited to one task. We are interested in showing --- via simulations --- that the proposed model performs well for a variety of tasks. We randomly generated instances of the reward matrix $R$ and $\alpha$ values and simulated runs of the partial and complete adaptation policies for increasing time horizons $T$. The simulated humans partially adapted to the robot, and the robot did not observe whether they learned. For each value of $T$, we randomly sampled 1000 reward matrices $R$ and simulated 100 policy runs for each sampled instance of $R$. Fig.~\ref{fig:performance}~(right) shows the reward averaged over the number of rounds $T$, policy runs and instances of $R$. For $T=1$, the mean reward is the same for both models, since there is no adaptation. The partial adaptation policies consistently outperform the complete adaptation ones for a large range of $T$. For large values of $T$ the performance difference decreases. This is because the human eventually learns the true payoffs and the gain from playing the true best response outweighs the initial loss caused by the complete adaptation model.

\noindent\textbf{Selection of $\alpha$.}
We note that the $\alpha$ value, which represents the probability that the human learns the true robot capabilities, is task and population-dependent. In our experiment, participants were recruited from a university campus, and most of them were able to infer that they should empty the bottle, after observing the robot failing and being notified that ``the robot exceeded its torque limits.'' Different participant groups may require a different $\alpha$ value. The value of $\alpha$ could also vary for different robot actions; we conjecture that our theoretical results hold also when there is a  different adaptation probability $\alpha_i$ for each row $i$ of the payoff matrix, which we leave as future work.



\section{Conclusion} 

We presented a game-theoretic model of human partial adaptation to the robot. The robot used this model to decide optimally between taking actions that reveal its capabilities to the human and taking the best action given the information that the human currently has. We proved that under certain observability assumptions, the optimal policy can be computed efficiently. Through a human subject experiment, we demonstrated that policies computed with the proposed model significantly improved human-robot team performance, compared to policies that assume complete adaptation of the human to the robot.

While our model assumes that the human may learn only the entries of the row played by the robot, there are cases where a robot action may affect entries that are associated with other actions, as well. For instance, Cha et al.~\cite{cha2015perceived} have shown that conversational speech can affect human perception of robot capability in physical tasks. We are excited to explore the structure of probabilistic graphical models of human adaptation, and use the theoretical insights from this work to develop efficient algorithms for the robot.

\section*{Acknowledgments}
{\small
 This work is funded by the DARPA SIMPLEX program through ARO contract number 67904LSDRP, the National Institutes of Health (\#R01EB019335), the National Science Foundation (CPS-1544797, CCF-1215883, IIS-1350598, CCF-1525932), the Office of Naval Research (N00014-16-1-3075), a Fulbright-Nehru postdoctoral fellowship, and a Sloan Research Fellowship. We also acknowledge the Onassis Foundation as a sponsor.
}

\bibliographystyle{IEEEtran}
\bibliography{mybib2short}  

\end{document}